\documentclass[sigconf,nonacm]{acmart}
\makeatletter
\@ifclasslater{acmart}{2026/05/31}{}{%
  \ClassError{acmart}{Loaded acmart predates style/acmart.cls (v2.18 2026/05/31).
    Build with latexmk from the repo root, or export TEXINPUTS=.:./style//:}{}}
\makeatother
\usepackage{xcolor}
\usepackage{capt-of}
\usepackage{amsmath}
\usepackage{float}

\usepackage{amssymb}
\AtBeginDocument{%
  }

\definecolor{darkred}{rgb}{0.6,0,0}
\definecolor{darkgreen}{rgb}{0,0.4,0}

\begin{document}
\title[Agentic Context Cracking: Token-Efficient Data Reasoning Agents via
Adaptive Structuring of Unstructured Data]{Agentic Context Cracking:
\texorpdfstring{\\}{ } Token-Efficient Data Reasoning Agents via Adaptive
Structuring of Unstructured Data}


\author{Milad Rezaei Hajidehi}
\affiliation{%
  \institution{Harvard University}
  \city{Cambridge}
  \state{Massachusetts}
  \country{USA}}
\email{milad@seas.harvard.edu}

\author{Qitong Wang}
\affiliation{%
  \institution{Harvard University}
  \city{Cambridge}
  \state{Massachusetts}
  \country{USA}}
\email{qitong@seas.harvard.edu}

\author{Stratos Idreos}
\affiliation{%
  \institution{Harvard University}
  \city{Cambridge}
  \state{Massachusetts}
  \country{USA}}
\email{stratos@seas.harvard.edu}

\begin{abstract}
Across open and enterprise settings, valuable data remains embedded in unstructured sources: web pages, reports, contracts, filings, earnings calls, and PDFs. The big bet in enterprise AI is deploying LLM agents that reason over this data to answer complex questions for every knowledge worker. Agents can do this today, but at prohibitive cost. Each question repeatedly opens large documents to recover scattered evidence, consuming up to a million tokens. However, if the data were already structured, the same question would reduce to a cheap database lookup. For example, on the FanOutQA benchmark, reasoning over an ideal pre-structured store is 28$\times$ cheaper, and the gap grows to orders of magnitude as questions fan out over more documents. Yet structuring everything in advance is not viable: documents hold vastly more possible structure than any workload will use, and the useful structure and documents are unknown until queries arrive.

We propose \textit{agentic context cracking}, an approach that
structures unstructured data adaptively and speculatively as a
byproduct of reasoning itself, in the spirit of database cracking and
adaptive data systems. Structuring is adaptive because observed
queries decide when it happens and what matters, and speculative
because it goes beyond the current question. Whenever the agent opens
a document to answer, a cracking sub-agent forks from the
already-loaded context at marginal cost and extracts grounded
structure likely to serve related future queries. Over time, an
increasing share of queries is fully covered by structured data and
answered without opening a document, keeping agentic accuracy at
close to RAG cost. On FanOutQA, extended with merely one related
question per test question, cracking cuts cost by 53\% while
preserving accuracy, and already runs 9$\times$ cheaper at the 10th
percentile of per-question savings. Agentic context cracking is a first
step toward next-generation data infrastructure for agentic reasoning
over unstructured data: a shared substrate beneath the model where
knowledge that reasoning already paid to uncover accumulates, so the
system improves with use for every agent and user that queries it.
\end{abstract}

\maketitle

\section{Introduction}
\label{sec:intro}
\begin{figure}[t]
  \centering
  \includegraphics[width=1\linewidth]{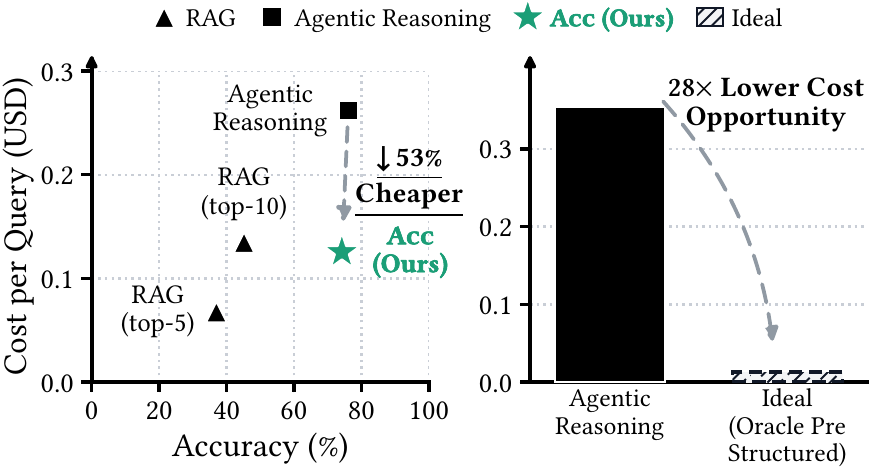}
  \caption{\underline{Left:} On the full FanOutQA~\cite{zhu-etal-2024-fanoutqa} benchmark over Wikipedia, agentic reasoning far outperforms fixed top-$k$ RAG but at high cost. Our approach, \textit{agentic context cracking}, preserves almost all of this accuracy advantage while reducing cost by 53\%. \underline{Right:} On ten sampled questions, we extracted every fact required to answer each question from the unstructured documents and stored it in a database. Reasoning over this ideal store is 28$\times$ cheaper. This gap is the cost reduction available to adaptive structuring systems.}
  \label{fig:1}
\end{figure}

\textbf{Knowledge and Insights Are Encased in Unstructured Data.}
Much of the world's knowledge is not stored in clean tables. It lives
in unstructured sources such as books, enterprise and legal
documents, web pages, SEC filings, earnings calls, and PDFs.
The evidence required to answer questions over these sources is
scattered across many documents. Consider this question: which
cast members worked with both Alfred Hitchcock and Martin Scorsese?
Over Wikipedia, answering it requires enumerating each director's
films from their pages, extracting cast members from
each film's page, connecting the resulting director-film and
film-cast relations, and intersecting the two cast sets. This flow
is general: at each step, evidence is gathered from documents,
assembled into intermediate entities, attributes, and relations, and
passed to later steps. The answer is not a set of retrieved passages.
It is a latent structured dataset, constructed step by step from
sparse evidence inside many large documents. We call this pattern
\textit{data reasoning}. 

\textbf{Data Reasoning over Unstructured Data is Already Everywhere
and Still Growing.} More than eighty commercial and open-source systems have publicly  launched to answer complex questions over the web and other large document collections~\cite{xu2025deepresearch, openai2025deepresearch, perplexity2025deepresearch}, and many more run privately inside enterprises~\cite{hebbia, glean}. Many of the questions real users ask follow exactly the data reasoning pattern~\cite{wolfson2026monaco}. 
Personal questions dominate in volume, and millions of people ask them over the open web. Professional questions carry higher stakes and target enterprise archives and public government records~\cite{opsahlong2026officeqapro}.
A financial analyst might ask which companies blamed supply chains on their earnings calls yet reported rising inventory in their annual reports, matching a single remark in an hour-long transcript against one line in a long filing. A health department might ask which suppliers appear in both food recall notices and restaurant inspection reports across multiple cities, with supplier names scattered through lengthy inspection narratives. Answering either means fanning out over hundreds of documents in multiple steps.

\textbf{Agentic Data Reasoning Extracts Encased Knowledge but at Prohibitive Cost.} Agentic systems can execute the multi-step flow needed to answer data reasoning questions. An agent plans, searches, opens documents, extracts facts, executes code, maintains intermediate state, and selects its next action based on the evidence gathered so far~\cite{chen2026agentir,asai2026openscholar}. These capabilities bring far higher answer quality than models relying on parametric knowledge or fixed top-$k$ retrieval~\cite{zhu-etal-2024-fanoutqa,opsahlong2026officeqapro}. Figure 1 (left) also shows this. This quality comes at high cost. Data reasoning is largely prefill-intensive: the agent repeatedly loads large documents into context to recover scattered facts, while the final answer is often short. Cost grows with dependent steps, entities, and documents. In our FanOutQA experiments, one question can consume up to one million tokens and cost nearly one dollar. Real-world data reasoning questions can have greater fan-out, evidence requirements, and reasoning depth, making them substantially more expensive~\cite{wolfson2026monaco}.

\textbf{Read it Once, Pay it Once: Queries Reveal Useful Structure.} 
If the data were magically structured, many of these questions, including our Hitchcock example, would reduce to a simple and low-cost Text2SQL query. We demonstrate this phenomenon in Figure~\ref{fig:1} (right). We sampled questions from FanOutQA, manually extracted every fact their answers require into a database, and reran the agent over the database instead of the raw documents. Each task reduced to data discovery and a short query, without opening a single document. This ideal pre-structured store makes reasoning $28\times$ cheaper. 
This factor is specific to FanOutQA, whose questions span seven documents on average~\cite{zhu-etal-2024-fanoutqa}. Once structure covers a question, the agent skips every document open, so the saving reaches orders of magnitude for real questions that fan out over hundreds of documents.
While LLMs are excellent at capturing such structure, structuring everything in advance is infeasible. Documents contain orders of magnitude more entities, attributes, and relations than any workload will use, and extracting them all is an unbounded decode. Moreover, we do not know in advance which structure a workload wants, or how often it will be reused. Only the workload reveals which structure is worth extracting.

Two insights turn this dead end into an opportunity. First, at the workload level, related queries reopen the same documents and repeat much of the same work, while each query reveals which structure and relations matter. 
This semantic locality is the norm, not the exception. From the user's side, conversations become investigations and users hold similar interests. From the data's side, the same documents get opened over and over because they hold many related facts and evidence that queries need. 
Second, at the inference level, once the agent loads a document, its prefill has already been computed and its KV-cache entries already exist. With careful prompt structure, shared-prefix serving~\cite{vllm23,sglang24}, and API prompt caching, a second generation over the same context avoids or sharply reduces repeated prefill cost, leaving decoded tokens as the main additional work. We give this decode a fixed budget that is small relative to the prefill it saves, yet sufficient to produce reusable structure. Together, the two insights suggest a move stronger than caching. Cached answers help only when the exact question returns. Speculating about useful structures goes much further: While the document is loaded and paid for, extract grounded evidence that related but distinct future queries will need.





\textbf{Agentic Context Cracking for Token/Cost-Efficient Data Reasoning.}
We introduce \textit{agentic context cracking}, an approach that brings the query-driven philosophy of database cracking and adaptive data systems to AI reasoning~\cite{itrails,idreos2007cracking}. The system never opens documents solely to extract structure. Instead, whenever the reasoning agent opens one to answer, a cracking sub-agent forks from the already-loaded context at low marginal cost. It speculates about grounded, evidence-backed entities, attributes, and relations likely to serve future queries. The system validates and catalogs these cracked objects. Later queries reuse them via structured reads, avoiding document opens entirely, and fall back to original documents on a miss. Unlike database cracking, which reorganizes relations, agentic context cracking creates useful structure where none exists.
This structuring is adaptive because observed queries guide what is worth extracting, and agentic because a reasoning sub-agent, not a fixed extractor or schema, uses semantic understanding to read each document and speculate about what to extract. \looseness=-1

The savings available to adaptive structuring are substantial. Longer-running workloads and smarter cracking unlock them. The ideal store is 28$\times$ cheaper on the FanOutQA benchmark, and this gap widens with fan-out. In our case study, which models an evolving investigation, cracking cuts cost by 3$\times$. On FanOutQA extended with only a single reuse chance per test question, one earlier related question, cracking still cuts cost by 53\% while preserving accuracy. At the tenth percentile of savings, cracking already runs 9$\times$ cheaper. Reuse separates the average from this best case, and reuse chances accumulate as the workload runs. The store an enterprise or AI service accumulates through use becomes a durable asset, a data moat that, unlike KV caches, transfers across models. \looseness=-1


\section{Agentic Reasoning: Capable but Expensive}
\label{sec:problem}

\textbf{Latent Data Reasoning Tasks.}
A major advance in language modeling has been the ability to decompose complex tasks into sequences of smaller reasoning steps.
Reasoning tasks take distinct forms: solving IOI/ICPC problems, constructing data science pipelines, and resolving knowledge-intensive questions over large unstructured corpora (e.g., the web, Wikipedia~\cite{zhu-etal-2024-fanoutqa}, and enterprise document collections~\cite{opsahlong2026officeqapro}).
In this paper, we focus on reasoning tasks that require an agent to extract knowledge from unstructured documents, construct intermediate data and supporting evidence at each step, and pass those artifacts to later steps. In these tasks, the answer comes from reasoning over the constructed intermediate data rather than retrieving a single fact. We call this class \textit{latent structured data reasoning}, or \textit{data reasoning} for brevity.


\textbf{Two Running Examples.} Throughout the paper, we use two
questions to explain the problem, intuitions, and solution:

(1) What is the career average rebounds per game for each NBA player
who has won the IBM Award?

(2) Who are all the cast members who ever worked with both Alfred
Hitchcock and Martin Scorsese?

The first question comes from FanOutQA~\cite{zhu-etal-2024-fanoutqa},
our primary benchmark. The second question is a harder data reasoning
task that we use in our case study. As illustrated in the
introduction, answering it requires a multi-step construction over
many Wikipedia pages. Although both questions appear casual, they
follow the same pattern as the high-stakes questions in the
introduction.

\textbf{Agentic Solutions: Capable but Expensive.}
 LLMs that rely only on parametric knowledge or conventional retrieval achieve low accuracy on data reasoning tasks~\cite{zhu-etal-2024-fanoutqa,opsahlong2026officeqapro}. Agentic systems instead imitate the data reasoning process by planning a sequence of steps and adaptively choosing when to invoke tools~\cite{chen2026agentir,asai2026openscholar}. The agent can search the corpus, open documents, use filesystem search/navigation API, extract facts, run python code, and store intermediate data for later steps. Rather than following a fixed pipeline, the agent selects its next action based on the question, the documents read so far, and the intermediate state. The agent can also manage its context by loading documents, extracting useful facts, and removing the documents after use. These capabilities substantially improve accuracy and grounding, but processing large documents makes each query expensive. For example, in our experiments with FanOutQA, 
 answering just a single question with the Haiku model consumes up to one million tokens and costs nearly a dollar, using Haiku Model and FanOutQA Agents~\cite{zhu-etal-2024-fanoutqa}. These costs limit the scalability and accessibility of agentic data reasoning.

\section{Agentic Context Cracking: Adaptive Structuring of Unstructured Data}
\label{sec:solution}

We show that data systems techniques can solve the cost bottleneck of data reasoning, substantially reducing token consumption. We show that the high cost of AI reasoning creates a pressing new class of problems for database research, requiring data systems and algorithms designed around data reasoning workloads. Prior work has likewise shown that data-management operators and abstractions improve unstructured-data processing~\cite{palimpzest,lotus,docetl,agentfirst,spear}. \looseness=-1

\textbf{Documents and Prefill Dominate Cost.}
Reasoning workloads have different inference profiles. In mathematical reasoning, little task-specific evidence is available, the models work over their own generated tokens; such workloads are often decode-intensive. Data reasoning has the opposite profile. Under an accuracy-first policy, the agent cannot speculate about missing facts. It must repeatedly open documents, load them into context, extract needed facts, and potentially remove them. Data reasoning is therefore largely prefill-intensive. This bottleneck intensifies with \textit{depth}, as reasoning spans dependent steps; \textit{width}, as each step fans out over many entities and documents; and \textit{sparsity}, as a few relevant tokens are scattered across much larger documents. In our cast-member example, the agent opens one page per film although scattered cast names occupy only a small fraction of each page.

\textbf{Avoiding Document Prefill Through Structure and Its Challenges.}
One way to avoid document opens is to magically structure document knowledge in a database. The prefill-heavy workload becomes a short SQL query without opening a document; for example, retrieving all cast members in Hitchcock films. Figure 1 (right) demonstrates the performance gap between raw documents and a pre-structured database. Yet extraction is difficult. Hand-designed algorithms over unstructured data are task-specific and potentially lossy. LLMs are more flexible but still require document prefill. Documents also contain many entities, attributes, and relations, making useful structure and its representation unclear. Extracting everything would require an unbounded decode. Unlike a database index, which organizes structure that already exists, exhaustive corpus extraction is infeasible and may never amortize.


\textbf{Adaptive Agentic Context Cracking: Queries Guide What to Extract.} 
We build our system to extract structure adaptively as queries reveal
what is useful over time. When the answer agent opens a document, we
fork a parallel cracking branch over it. The cracking sub-agent sees
current and prior queries over the document. Using semantic
reasoning, our instructions, and our data model, it speculates about
reusable entity sets, attributes, and relations. Crucially, it is not
limited to the attribute or entity set requested by the
query. It also extracts nearby structure likely to support related
queries. Auxiliary algorithms normalize and validate the output
before storage. Extraction is semantic, selective, and
guided by observed demand. This design makes sense from both
application and low-level LLM inference perspectives.

\begin{figure}[t]
  \centering
  \includegraphics[width=\linewidth]{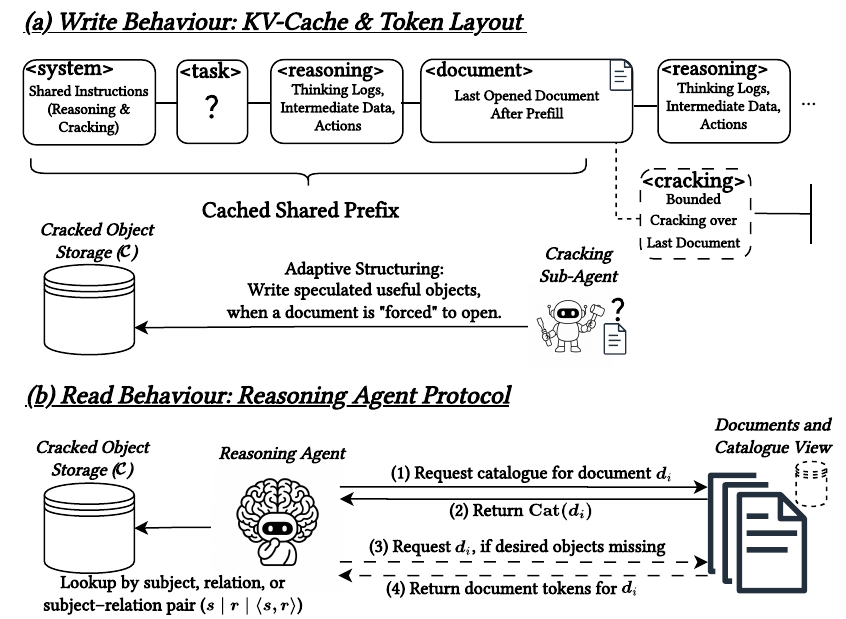}
  \caption{(a) The cracking branch decodes over a shared prefix to populate cracked-object store for later queries. (b) Reasoning protocol with catalogue lookup and document fallback.}
  \label{fig:sol}
\end{figure}

From the application perspective, the approach is effective when queries seek related objectives over overlapping entity sets. When a query opens NBA player pages to obtain career points per game for all-time scorers, the cracking sub-agent can also extract career rebounds per game. Our running question can then reuse those values for overlapping IBM Award winners without reopening their pages. Such locality grows as conversations become investigations and organizations form recurring categories and document hotspots around the same companies, products, cases, and people. From the performance perspective, the document’s KV cache already exists when cracking begins. We reuse the cached prefix and fork a second generation under a cracking instruction (Figure~\ref{fig:sol}(a)), avoiding another prefill through shared-prefix KV-cache reuse~\cite{sglang24,vllm23}. The cracking branch runs in parallel and outside the answer path, so it does not delay the current query. Its only additional model work is the bounded decode that produces the structured output.

We call this approach \textit{agentic context cracking} (\textsc{Acc}), \textit{context} because the sub-agent cracks documents already loaded and paid for in the reasoning agent’s context, and \textit{cracking} because it shares database cracking’s central principle: observed queries drive adaptive, incremental organization for future reuse. However, unlike database cracking, which reorganizes existing relations, agentic context cracking adaptively creates useful structure where none exists. Adaptation is therefore more consequential here: exhaustive extraction is infeasible, KV-cache reuse keeps overhead minor and justifiable, and the payoff is larger LLM inference savings. It remains durable and usable across models.

\newcommand{\CrackedObject}{c}
\newcommand{\CrackedObjects}{\mathcal{C}}

\newcommand{\Subject}{s}
\newcommand{\RelationLabel}{r}
\newcommand{\Object}{o}
\newcommand{\Cardinality}{\kappa}
\newcommand{\ObjectUnit}{u}
\newcommand{\Evidence}{\varepsilon}
\newcommand{\Document}{d}
\newcommand{\Region}{\rho}

\newcommand{\EntityDomain}{\mathcal{E}}
\newcommand{\RelationDomain}{\mathcal{R}}
\newcommand{\ValueDomain}{\mathcal{V}}
\newcommand{\StringDomain}{\Sigma^{*}}
\newcommand{\IntegerDomain}{\mathbb{Z}}
\newcommand{\DateDomain}{\mathbb{D}}
\newcommand{\DocumentDomain}{\mathcal{D}}
\newcommand{\UnitDomain}{\mathcal{U}}
\newcommand{\CardinalityDomain}{\mathcal{K}}

\newcommand{\StringTag}{\mathsf{str}}
\newcommand{\IntegerTag}{\mathsf{int}}
\newcommand{\DateTag}{\mathsf{date}}
\newcommand{\Singular}{\mathsf{singular}}
\newcommand{\ListCardinality}{\mathsf{list}}
\newcommand{\NoUnit}{\bot}

\newcommand{\DefinedAs}{\triangleq}

\newcommand{\CrackedTuple}[6]{%
  \left\langle #1,#2,#3,#4,#5,#6 \right\rangle}

\newcommand{\EvidenceTuple}[2]{%
  \left\langle #1,#2 \right\rangle}

\newcommand{\DocCatalogue}[1]{%
  \mathsf{Cat}\!\left(#1\right)}

\newcommand{\EvidenceDocument}[1]{%
  \operatorname{doc}\!\left(#1\right)}

\textbf{Cracked Objects \& Data Model.}
Our system operates over facts extracted by a cracking sub-agent running
on an inference branch forked from the answer agent after a document
enters context. We call these facts \textit{cracked objects}. We model
each cracked object as an extended RDF-style edge annotated with
cardinality, unit, and evidence:
\begingroup
\setlength{\abovedisplayskip}{4pt}
\setlength{\belowdisplayskip}{4pt}
\setlength{\abovedisplayshortskip}{4pt}
\setlength{\belowdisplayshortskip}{4pt}
\setlength{\jot}{1pt}
$$
\begin{aligned}
\CrackedObject
&=
\CrackedTuple
  {\Subject}
  {\RelationLabel}
  {\Object}
  {\Cardinality}
  {\ObjectUnit}
  {\Evidence}
\in\CrackedObjects,
\qquad
\Subject\in\EntityDomain,\quad
\RelationLabel\in\RelationDomain,\quad
\Object\in\EntityDomain\cup\ValueDomain,
\\
\ValueDomain
&\DefinedAs
\bigl(\{\StringTag\}\times\StringDomain\bigr)
\cup
\bigl(\{\IntegerTag\}\times\IntegerDomain\bigr)
\cup
\bigl(\{\DateTag\}\times\DateDomain\bigr).
\end{aligned}
$$
\endgroup
Here, $\EntityDomain$ and $\RelationDomain$ are open domains populated
by the cracking sub-agent. An object ($\Object$) is either an entity or a tagged
scalar, which pairs an explicit type with a value. The current schema
supports strings, integers, and sortable dates. A cracked object
therefore represents an entity-to-entity or entity-to-value relation.

Each cracked object carries evidence
$\Evidence=\EvidenceTuple{\Document}{\Region}$, where
$\Document\in\DocumentDomain$ identifies its source and
$\Region$ identifies the supporting region within that source. The cardinality
$\Cardinality\in
\CardinalityDomain=
\{\Singular,\ListCardinality\}$
distinguishes single-object relations from relations that yield complete
sets. It defines completeness during reuse: one edge completes a
singular relation, whereas a list relation is reusable only after all
members are extracted. We store list members as separate edges so each
remains independently indexable and joinable. The optional
$\ObjectUnit\in\UnitDomain\cup\{\NoUnit\}$ records a canonical unit. In our model, documents are logical units that may represent a file, revision, or chunk.

\textbf{Read/Reasoning Behaviour: Using Cracked Objects.}
Like the conventional baseline agent, our main reasoning agent has
filesystem-style tools for searching, navigating, and opening documents.
However, it also benefits from new constrained tool-calls that support
structured reads by subject, relation, and subject--relation pair from
the cracked-object store. These calls let the agent reuse compact
cracked objects without opening or prefilling raw documents. When
entity-name or relation-label mismatches occur, they prevent retrieval
of relevant stored objects. We therefore introduce the
\textit{catalogue}, a logical view over the cracked-object store. For
each document $\Document$, the catalogue lists available
subject--relation pairs with their canonical labels and cardinalities:
\begingroup
\setlength{\abovedisplayskip}{0pt}
\setlength{\belowdisplayskip}{1pt}
\setlength{\abovedisplayshortskip}{0pt}
\setlength{\belowdisplayshortskip}{1pt}
$$
\DocCatalogue{\Document}
\DefinedAs
\pi_{\Subject,\RelationLabel,\Cardinality}
\left(
  \sigma_{\EvidenceDocument{\Evidence}=\Document}
  (\CrackedObjects)
\right).
$$
\endgroup
The catalogue helps the agent resolve available subjects and relations
before issuing a structured read (Figure~\ref{fig:sol}(b)). The system surfaces it with the
document search result before the agent decides whether to open the
document. If the desired structure is present, the agent issues the
corresponding structured read and receives compact values with evidence,
avoiding document prefill. Otherwise, it opens and prefills the raw
document. A missing structured result means that the required structure
is unavailable or unresolved. Raw-document access remains the
conservative fallback.

\begin{figure*}[t]
  \centering
  \includegraphics[width=1\linewidth]{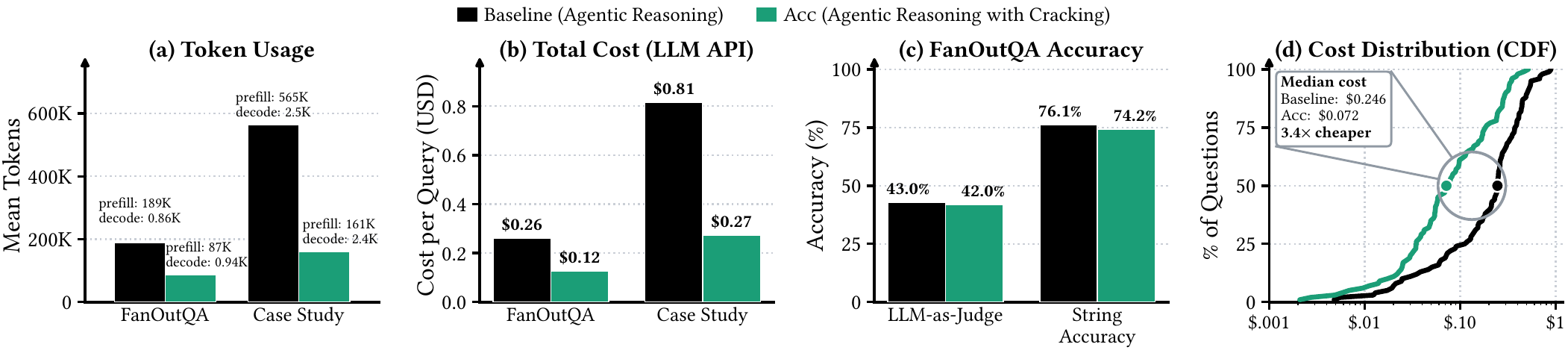}
  \caption{Performance of Agentic Data Reasoning with and without Context Cracking.}
  \label{fig:3}
\end{figure*}

\textbf{Write/Cracking Behaviour: Producing Cracked Objects.}
When cracked-object reads cannot satisfy an information need, the main reasoning agent resolves the resulting ``miss'' by opening the raw document. The system never opens documents solely for cracking. Instead, after the document enters context, a cracking sub-agent forks from the reasoning branch and targets that document. In local serving, this fork can reuse the document's KV-cache prefix, avoiding a second prefill. For API models with prompt caching, the shared prefix hits the cache and is billed at a lower rate. Given the document and missed information need, the sub-agent does not merely cache the requested fact or extract predefined entities and attributes. It uses semantic reasoning to speculate about which additional entities, attributes, and relations may support future queries, while extracting only grounded facts. 

In our NBA example, a miss on one player statistic may prompt extraction of other
reusable player statistics. Cracking thus resembles prefetching after a
cache miss, but replaces fixed address-based locality with
agent-discovered semantic locality. Under a fixed output-token budget,
the sub-agent emits schema-constrained JSON. It emits a complete
list-valued relation or omits it. Post-processing validates and
normalizes the output, expands lists into separate edges, and inserts
only grounded cracked objects. Because cracking remains outside the
answer dependency path, it neither blocks the current answer nor
prevents the reasoning agent from using the raw document if cracking
fails. 
The read and write behaviour capture the core mechanism. The
implementation also includes cracking instructions and invariant checks.
For instance, the system canonicalizes numbers, units, and dates and
requires each invocation to crack the entire logical document, avoiding
the complexity and ambiguity of partially cracked data.

\section{Evaluation}
\label{sec:evaluation}

We test if agentic context cracking captures reusable structure across related questions with related but distinct  objectives and overlapping but distinct entity sets. We measure how much this reuse reduces inference cost and whether it preserves answer quality. \looseness-1

\textbf{Evaluation Setup.} The baseline and our system are built on Claude-Haiku-4.5 with FanOutQA's specifications, instructions, prompts, and tool calls to access Wikipedia and the filesystem. Our system differs from the baseline only in the cracking interface. It adds cracking instructions and cracked-object read/write calls. Prompt caching was enabled during all experiments. We measure prefill and decode tokens, API cost, and answer accuracy.

Our benchmarks are FanOutQA~\cite{zhu-etal-2024-fanoutqa} and a cinema case study centered on Alfred Hitchcock. FanOutQA contains wide, multi-hop questions over Wikipedia, ranging from Supreme Court justices to NBA players. Its questions almost never revisit the same documents and therefore lack the locality our system targets. This locality is absent from reasoning benchmarks by design, since they assess the quality of individual queries. Real-world workloads instead revisit documents through related queries.

To recover this locality without altering the test questions, we double the workload: for each test question, an LLM (claude-opus-5) generates one related but distinct question, and a human verifies it. Each generated question targets a different attribute over an overlapping entity set. The generator sees only the target question and our criteria: answerable from Wikipedia, overlapping entities, and a distinct attribute. It is unaware of the cracking system, its design, implementation, or outputs. For our running NBA example, the generated question asks for the career points per game of the top ten all-time NBA scorers. Caching answers cannot help here. No generated question repeats or paraphrases its test question, and the two never share the requested attribute. Any saving comes from semantic speculation about related attributes. In both FanOutQA and the case study, all primary metrics are reported on the original test questions only. First, this keeps the FanOutQA accuracy evaluation well defined because the generated related questions have no gold answers. Second, the cracked-object store is populated entirely by the preceding related questions, allowing us to evaluate a warmed store without the noise of an empty-store cold start. We report the additional cracking cost separately.

The case study models a longer investigation, closer to cracking in long-running systems. It has fewer test questions than FanOutQA but more related questions, around a single interest. We first ask 20 questions about Hitchcock's films, career, and life. We evaluate 10 related test questions that reuse these cracked objects. The sequence represents an evolving film investigation.

\textbf{Results.} Figure 1 (left) is our headline. Cracking keeps agentic accuracy at close to RAG cost. All systems, including the RAG baselines, use the same model. Figure 1 (right) shows the opportunity for adaptive structuring systems. With either ideal cracking or longer-running systems, cost converges toward the ideal structured store. Figure~\ref{fig:3} breaks the headline down. Figure~\ref{fig:3}(a) shows token usage. Cracking cuts mean prefill from 189K to 87K tokens on FanOutQA and from 565K to 161K on the case study. Decode stays below 3K tokens in all settings. We also report cost as a comprehensive metric and the most important one for AI applications. Figure~\ref{fig:3}(b) shows that mean cost per question drops from \$0.26 to \$0.12 on FanOutQA and from \$0.81 to \$0.27 on the case study. The case study is more expensive overall because its questions are harder, with two to four times the fan-out. Its improvement is also larger because twenty related questions come first, so test questions are more likely to find cracked objects to reuse. Figure~\ref{fig:3}(c) shows accuracy. 
Both metrics come from FanOutQA. String accuracy is the fraction of gold-answer facts matched at the word level. LLM-as-judge scores the answer against the gold answer and instructions using GPT, as specified by the benchmark. The judge is harsher on partially correct answers. LLM-judged accuracy differs by one point from the baseline (42\% vs. 43\%, $p$-value $= 0.39$), with no statistically significant difference and word-match accuracy drops by two points, mostly due to answer canonicalization. The similar accuracy is rooted in our design where we prioritize a robust fallback. When reuse misses, the agent reads the original document. A more aggressive policy could trade this robustness for larger savings.

Figure~\ref{fig:3}(d) shows the independently sorted per-question cost distributions. The median question costs \$0.246 under the baseline and \$0.072 under cracking, a 3.4$\times$ ratio of medians. The gain is not uniform. For paired question-by-question cost ratios, cracking is 9$\times$ cheaper at the 10th percentile and costs 1.24$\times$ the baseline at the 90th percentile. These are questions with no reuse, where cracking and data discovery add overhead without savings. Cracking pays off on three quarters of the questions and wins clearly in aggregate. 

We set a 4K-token cracking decode budget per question, spent across cracking forks. On FanOutQA, this adds 12\% overhead per question. It includes the decode cost and cache read that vendors charge for previous context when forking. This budget produces roughly 150 cracked objects. Larger budgets yield proportionally more objects, but the additional objects tend to be narrower and less likely to be reused. This cost amortizes once a question's cracked objects let later questions avoid one or two document opens. A few avoided opens repay the budget, and every open beyond that saves a document prefill. Smarter budget allocation (e.g., per specific query or document) remains future work. Even with our simple allocation and cracking overhead, the gains in Figure~\ref{fig:3} hold by a wide margin. \looseness-1

\section{Related Work}
\label{sec:related}

Deep-research and reasoning agents answer complex questions over large corpora with
multi-step search, document opens, and evidence
synthesis~\cite{asai2026openscholar,chen2026agentir}.
Prior work introduces data-management operators, abstractions, and
execution strategies for processing unstructured
data~\cite{palimpzest,lotus,docetl,agentfirst,spear,russo2025deepresearch}. Agentic context cracking
complements both lines by speculating and accumulating reusable, evidence-backed structure that
these agents and operators can run over, so later queries reuse the document
opens that earlier queries paid for.

OpenIE mines relational tuples from text without a predefined schema~\cite{etzioni2008open}. Knowledge-base construction, knowledge graphs, and graph-based RAG build representations before queries arrive~\cite{zhang2017deepdive,vrandecic2014wikidata,edge2024graphrag}. Query-driven document analytics extracts only values needed by the current query~\cite{lin2024zendb}. Dataspaces add integration structure incrementally as queries reveal what matters~\cite{itrails}. Our approach sits between upfront construction and per-query extraction. The data model is open, queries guide extraction, and cracking builds structure incrementally from documents already opened by queries. Semantic caches return answers to near-duplicate questions~\cite{bang2023gptcache}. Cracked objects remain compact, evidence-backed, and reusable across distinct queries. Serving systems persist document and shared-prefix KV state to skip prefill~\cite{gim2024promptcache,liu2024cachegen,yao2025cacheblend,qin2025mooncake}. Cached KV state stays bound to one model, is far larger than its source text, occupies GPU or host memory, and adds transfer and cache-read costs. Cracking complements KV-cache reuse by reading the cached prefix. Cracked structures are plain text, so they survive model upgrades, transfer across LLMs, and compound into an organizational asset that continues saving tokens and cost.

Agent memory systems persist information across sessions \cite{packer2023memgpt, chhikara2025mem0, zhong2024memorybank, xu2025amem, rasmussen2025zep, letta}.
One can also view agentic context cracking as a memory system,
but a memory with \textit{speculation} and \textit{prefetching}: it memorizes and speculates structure over
the corpus rather than a user's preferences and past answers, and it writes
that structure, amortizing extraction cost across future
queries.
MemGPT pages conversation history between a bounded context window and
external storage~\cite{packer2023memgpt}, and Letta extends that design into
an agent framework~\cite{letta}. Mem0 extracts and consolidates salient facts
from ongoing conversations into a long-term store~\cite{chhikara2025mem0},
MemoryBank decays and reinforces stored memories over
time~\cite{zhong2024memorybank}, Zep maintains a temporal knowledge graph over
conversational and business data~\cite{rasmussen2025zep}, and A-Mem organizes
memories as dynamically linked notes~\cite{xu2025amem}.
Memory writes in these systems are triggered by conversation
turns and scoped to what was discussed.
\section{Discussion}
\textbf{Design Decisions, Limitations, and Future Work.} We expose constrained read functions rather than the full SQL space. This keeps the tool interface small, fallbacks explicit, and generated queries reliable. However, SQL supports counting, aggregation, joins, and direct multi-hop queries. Safely exposing these capabilities remains interesting future work. Our current system assumes a static corpus. Edits could be handled by dropping the affected cracked objects or incrementally refining them from the changes.

Our benchmarks use open, commonly trained-on Wikipedia content. Private office documents are unseen by the model and organization-specific. They are also repeatedly queried across users and workflows, making them an even more interesting workload for cross-query reuse. To reflect real-world workloads, we need benchmarks that capture the long history of queries over a document, ideally drawn from deployed systems. Such benchmarks would expose a large and practical opportunity to improve both individual queries and long-running real-world systems.
\section{Conclusion}
Agentic data reasoning can answer complex questions over unstructured corpora, but repeatedly reading documents makes it expensive. We introduced agentic context cracking. It uses observed queries to speculate about useful structure and extract it from documents already opened during reasoning. Future queries can then reuse this structure. The design exploits locality twice: related queries reuse extracted structure, and cracking itself reuses the document's KV cache. We showed that agentic context cracking reduces token usage and API cost while preserving answer quality. The remaining gap to an ideal fully structured store suggests a broad opportunity for data systems to improve agentic reasoning, especially in workloads where useful structure emerges across queries. This direction connects decades of work on the Semantic Web, entity resolution, and adaptive data systems with the reasoning ability of modern agents. Combining these foundations can make agentic reasoning more scalable and reliable. Agentic context cracking is an initial step toward such adaptive data systems for AI reasoning. Beyond our system, the idea applies wherever systems repeatedly reason over large unstructured documents.

\bibliographystyle{style/ACM-Reference-Format}
\bibliography{bib/sample-base}


\end{document}